\documentclass[10pt,twocolumn,letterpaper]{article}

\usepackage{cvpr}
\usepackage{times}
\usepackage{epsfig}
\usepackage{graphicx}
\usepackage{amsmath}
\usepackage{amssymb}
\usepackage{multirow}
\usepackage{bm}
\usepackage{indentfirst}

\usepackage[breaklinks=true,bookmarks=false]{hyperref}

\cvprfinalcopy 

\def\cvprPaperID{3727} 

\ifcvprfinal\fi
\begin{document}
\vspace{-16pt}

\title{CSRNet: Dilated Convolutional Neural Networks for Understanding the Highly Congested Scenes\vspace{-16pt}}
\vspace{-16pt}

\author{Yuhong Li$^{1,2}$, Xiaofan Zhang$^{1}$, Deming Chen$^{1}$ \\
        $^{1}$University of Illinois at Urbana-Champaign\\
        $^{2}$Beijing University of Posts and Telecommunications\\
        \large \textit{\{leeyh,xiaofan3,dchen\}@illinois.edu \vspace{-16pt}}
}

\maketitle
\thispagestyle{empty}

\begin{abstract}
\vspace{-8pt}

We propose a network for Congested Scene Recognition called CSRNet to provide a data-driven and deep learning method that can understand highly congested scenes and perform accurate count estimation as well as present high-quality density maps. The proposed CSRNet is composed of two major components: a convolutional neural network (CNN) as the front-end for 2D feature extraction and a dilated CNN for the back-end, which uses dilated kernels to deliver larger reception fields and to replace pooling operations. CSRNet is an easy-trained model because of its pure convolutional structure. We demonstrate CSRNet on four datasets (ShanghaiTech dataset, the UCF\_CC\_50 dataset, the WorldEXPO'10 dataset, and the UCSD dataset) and we deliver the state-of-the-art performance. In the ShanghaiTech Part\_B dataset, CSRNet achieves  47.3\% lower Mean Absolute Error (MAE) than the previous state-of-the-art method. We extend the targeted applications for counting other objects, such as the vehicle in TRANCOS dataset. Results show that CSRNet significantly improves the output quality with 15.4\% lower MAE than the previous state-of-the-art approach. 
\end{abstract}

\vspace{-8pt}
\section{Introduction}
\label{sec: Intro}

\begin{figure}[t]
\begin{center}
\includegraphics[width=\linewidth]{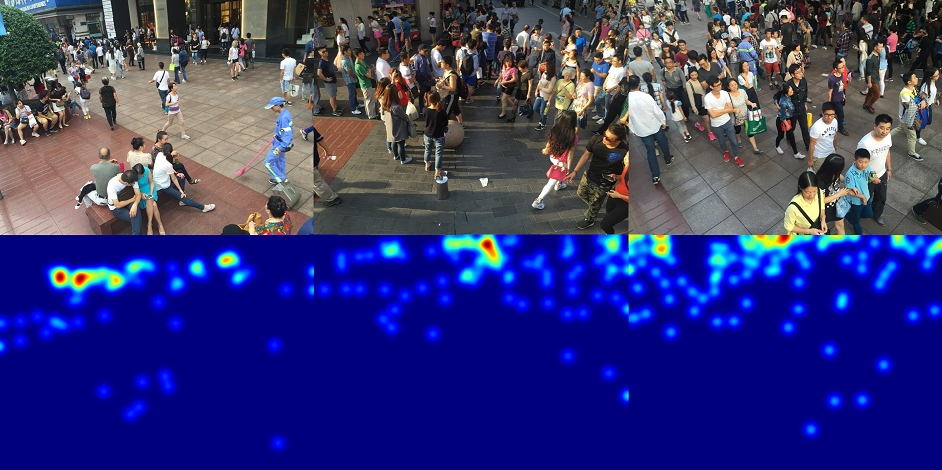}
\end{center}
   \caption{Pictures in first row show three images all containing 95 people in ShanghaiTech Part\_B dataset~\cite{zhang2016single}, while having totally different spatial distributions. Pictures in second row show their density maps.}
\label{fig:1}
\vspace{-10pt}
\end{figure}

Growing number of network models have been developed~\cite{zhan2008crowd,li2015crowded,zhang2015cross,sam2017switching,sindagi2017generating} to deliver promising solutions for crowd flows monitoring, assembly controlling, and other security services. 
%
Current methods for congested scenes analysis are developed from simple crowd counting (which outputs the number of people in the targeted image) to density map presenting (which displays characteristics of crowd distribution) ~\cite{zhang2016data}. This development follows the demand of real-life applications since the same number of people could have completely different crowd distributions (as shown in Fig.~\ref{fig:1}), so that just counting the number of crowds is not enough. The distribution map helps us for getting more accurate and comprehensive information, which could be critical for making correct decisions in high-risk environments, such as stampede and riot. 
However, it is challenging to generate accurate distribution patterns. One major difficulty comes from the prediction manner: since the generated density values follow the pixel-by-pixel prediction, output density maps must include spatial coherence so that they can present the smooth transition between nearest pixels.
Also, the diversified scenes, e.g., irregular crowd clusters and different camera perspectives, would make the task difficult, especially for using traditional methods without deep neural networks (DNNs). 
The recent development of congested scene analysis relays on DNN-based methods because of the high accuracy they have achieved in semantic segmentation tasks~\cite{long2015fully,wei2017stc,wei2017object,yu2015multi,chen2016deeplab} and the significant progress they have made in visual saliency~\cite{pan2016shallow}. The additional bonus of using DNNs comes from the enthusiastic hardware community where DNNs are rapidly investigated and implemented on GPUs~\cite{jia2014caffe}, FPGAs~\cite{Qiu:2016:GDE:2847263.2847265,zhang2017high,zhang2017machine}, and ASICs~\cite{andri2016yodann}. Among them, the low-power, small-size schemes are especially suitable for deploying congested scene analysis in surveillance devices. 

\begin{figure}[!t]
\begin{center}
\includegraphics[width=\linewidth]{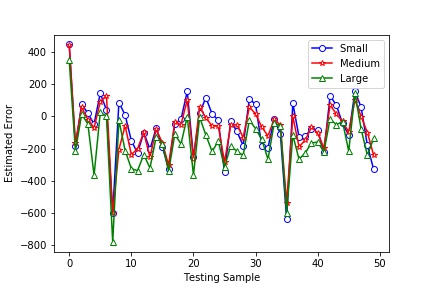}
\end{center}
\vspace{-10pt}
   \caption{The estimated error of 50 samples from the testing set in ShanghaiTech Part\_A~\cite{zhang2016single} generated by the three pre-trained columns of MCNN. Small, Medium, Large respectively stand for the columns with small, medium or large kernels in the MCNN.}
\label{fig:sml}

\end{figure}

Previous works for congested scene analysis are mostly based on multi-scale architectures~\cite{sam2017switching,sindagi2017generating,zhang2016single,boominathan2016crowdnet,onoro2016towards}. They have achieved high performance in this field but the designs they used also introduce two significant disadvantages when networks go deeper: large amount of training time and non-effective branch structure (e.g., multi-column CNN (MCNN) in~\cite{zhang2016single}).
We design an experiment to demonstrate that the MCNN does not perform better compared to a deeper, regular network in Table ~\ref{tab:1}. The main reason of using MCNN in ~\cite{zhang2016single} is the flexible receptive fields provided by convolutional filters with different sizes across the column. Intuitively, each column of MCNN is dedicated to a certain level of congested scene. However, the effectiveness of using MCNN may not be prominent. We present Fig.~\ref{fig:sml} to illustrate the features learned by three separated columns (representing large, medium, and small receptive fields) in MCNN and evaluate them with ShanghaiTech Part\_A~\cite{zhang2016single} dataset. The three curves in this figure share very similar patterns (estimated error rate) for 50 test cases with different congest densities meaning that each column in such branch structure learn nearly identical features. It performs against the original intention of the MCNN design for learning different features for each column. 

In this paper, we design a deeper network called CSRNet for counting crowd and generating high-quality density maps. Unlike the latest works such as~\cite{sam2017switching,sindagi2017generating} which use the deep CNN for ancillary, we focus on designing a CNN-based density map generator. Our model uses pure convolutional layers as the backbone to support input images with flexible resolutions. To limit the network complexity, we use the small size of convolution filters (like \begin{math}3\times 3\end{math}) in all layers. We deploy the first 10 layers from VGG-16~\cite{simonyan2014very} as the front-end and dilated convolution layers as the back-end to enlarge receptive fields and extract deeper features without losing resolutions (since pooling layers are not used). 
By taking advantage of such innovative structure, we outperform the state-of-the-art crowd counting solutions (a MCNN based solution called CP-CNN~\cite{sindagi2017generating}) with 7\%, 47.3\%, 10.0\%, and 2.9\% lower Mean Absolute Error (MAE) in ShanghaiTech~\cite{zhang2016single} Part\_A, Part\_B, UCF\_CC\_50~\cite{idrees2013multi}, and WorldExpo'10~\cite{zhang2015cross} datasets respectively. Also, we achieve high performance on the UCSD dataset~\cite{4587569} with 1.16 MAE. After extending this work to vehicle counting on TRANCOS dataset~\cite{onoro2016towards}, we achieve 15.4\% lower MAE than the current best approach, called FCN-HA~\cite{DBLP:journals/corr/ZhangWCM17aa}.

The rest of the paper is structured as follows. Sec.~\ref{sec:Related work} presents the previous works for crowd counting and density map generation. Sec.~\ref{sec:Proposed method} introduces the architecture and configuration of our model while Sec.~\ref{sec:Experiments} presents the experimental results on several datasets. In Sec.~\ref{sec:Conclusion}, we conclude the paper.

\begin{table}
\label{tab:1}
\begin{center}
\begin{tabular}{|l|c|c|c|}
\hline
Method & Parameters & MAE & MSE\\
\hline\hline
Col. 1 of MCNN
 & 57.75k & 141.2& 206.8 \\
\hline
Col. 2 of MCNN
 & 45.99k & 160.5& 239.0 \\
\hline
Col. 3 of MCNN
 & 25.14k & 153.7& 230.2\\
\hline
MCNN Total & 127.68k & 110.2& 185.9\\
\hline
A deeper CNN&83.84k
 & \textbf{93.0} & \textbf{142.2}\\
\hline
\end{tabular}
\end{center}
\caption{To demonstrate that MCNN~\cite{zhang2016single} may not be the best choice, we design a deeper, single-column network with fewer parameters compared to MCNN. The architecture of the proposed small network is: $CR(32,3)-M-CR(64,3)-M-CR(64,3)-M-CR(32,3)-CR(32,3)-CR(1,1)$. $CR(m,n)$ represents the convolutional layer with m filters whose size is $n\times n$ followed by the ReLu layer. $M$ is the max-pooling layer.  Results show that the single-column version achieves higher performance on ShanghaiTech Part\_A dataset~\cite{zhang2016single} with the lowest MAE and Mean Squared Error (MSE)\vspace{-4pt}}
\vspace{-10pt}
\end{table}

\section{Related work}
\label{sec:Related work}
Following the idea proposed by Loy \etal.~\cite{loy2013crowd}, the potential solutions for crowd scenes analysis can be classified into three categories:  detection-based methods, regression-based methods, and density estimation-based methods. By combining the deep learning, the CNN-based solutions show even stronger ability in this task and outperform the traditional methods.

\subsection{Detection-based approaches}
Most of the early researches focus on detection-based approaches using a moving-window-like detector to detect people and count their number~\cite{dollar2012pedestrian}. These methods require well-trained classifiers to extract low-level features from the whole human body (like Haar wavelets~\cite{viola2004robust} and HOG (histogram oriented gradients)~\cite{dalal2005histograms}). However, they perform poorly on highly congested scenes since most of the targeted objects are obscured. To tackle this problem, researchers detect particular body parts instead of the whole body to complete crowd scenes analysis~\cite{felzenszwalb2010object}.

\subsection{Regression-based approaches}
Since detection-based approaches can not be adapted to highly congested scenes, researchers try to deploy regression-based approaches to learn the relations among extracted features from cropped image patches, and then calculate the number of particular objects. More features, such as foreground and texture features, have been used for generating low-level information~\cite{chan2009bayesian}. Following similar approaches, Idrees \etal.~\cite{idrees2013multi} propose a model to extract features by employing Fourier analysis and SIFT (Scale-invariant feature transform)~\cite{790410} interest-point based counting.

\subsection{Density estimation-based approaches}
When executing the regression-based solution, one critical feature, called saliency, is overlooked which causes inaccurate results in local regions. 
Lempitsky \etal.~\cite{lempitsky2010learning} propose a method to solve this problem by learning a linear mapping between features in the local region and its object density maps. It integrates the information of saliency during the learning process. Since the ideal linear mapping is hard to obtain, Pham \etal.~\cite{pham2015count} use random forest regression to learn a non-linear mapping instead of the linear one.

\subsection{CNN-based approaches}
Literature also focuses on the CNN-based approaches to predict the density map because of its success in classification and recognition~\cite{krizhevsky2012imagenet,simonyan2014very,chollet2016xception}. 
In the work presented by Walach and Wolf~\cite{walach2016learning}, a method is demonstrated with selective sampling and layered boosting. Instead of using patch-based training, Shang \etal.~\cite{shang2016end} try an end-to-end regression method using CNNs which takes the entire image as input and directly outputs the final crowd count.
Boominathan \etal.~\cite{boominathan2016crowdnet} present the first work purely using convolutional networks and dual-column architecture for generating density map. Marsden \etal.~\cite{marsden2016fully} explore single-column fully convolutional networks while Sindagi \etal.~\cite{sindagi2017cnn} propose a CNN which uses the high-level prior information to boost the density prediction performance. An improved structure is proposed by Zhang \etal.~\cite{zhang2016single} who introduce a multi-column based architecture (MCNN) for crowd counting. Similar idea is shown in Onoro and Sastre~\cite{onoro2016towards} where a scale-aware, multi-column counting model called Hydra CNN is presented for object density estimation. It is clear that the CNN-based solutions outperform the previous works mentioned in Sec. 2.1 to 2.3.

\subsection{Limitations of the state-of-the-art approaches}

Most recently, Sam \etal.~\cite{sam2017switching} propose the Switch-CNN using a density level classifier to choose different regressors for particular input patches. Sindagi \etal.~\cite{sindagi2017generating} present a Contextual Pyramid CNN, which uses CNN networks to estimate context at various levels for achieving lower count error and better quality density maps. 
These two solutions achieve the state-of-the-art performance, and both of them used multi-column based architecture (MCNN) and density level classifier. However, we observe several disadvantages in these approaches: (1) Multi-column CNNs are hard to train according to the training method described in work~\cite{zhang2016single}. Such bloated network structure requires more time to train.  (2) Multi-column CNNs introduce redundant structure as we mentioned in Sec.~\ref{sec: Intro}. Different columns seem to perform similarly without obvious differences. (3) Both solutions require density level classifier before sending pictures in the MCNN. However, the granularity of density level is hard to define in real-time congested scene analysis since the number of objects keeps changing with a large range. Also, using a fine-grained classifier means more columns need to be implemented which makes the design more complicated and causes more redundancy. 
(4) These works spend a large portion of parameters for density level classification to label the input regions instead of allocating parameters to the final density map generation. Since the branch structure in MCNN is not efficient, the lack of parameters for generating density map lowers the final accuracy. Taking all above disadvantages into consideration, we propose a novel approach to concentrate on encoding the deeper features in congested scenes and generating high-quality density map. 

\section{Proposed Solution}
\label{sec:Proposed method}

The fundamental idea of the proposed design is to deploy a deeper CNN for capturing high-level features with larger receptive fields and generating high-quality density maps without brutally expanding network complexity. In this section, we first introduce the architecture we proposed, and then we present the corresponding training methods.

\subsection{CSRNet architecture}
Following the similar idea in ~\cite{boominathan2016crowdnet,sam2017switching,sindagi2017generating}, we choose VGG-16~\cite{simonyan2014very} as the front-end of CSRNet because of its strong transfer learning ability and its flexible architecture for easily concatenating the back-end for density map generation.   
In CrowdNet~\cite{boominathan2016crowdnet}, the authors directly carve the first 13 layers from VGG-16 and add a \begin{math}1\times 1\end{math} convolutional layer as output layer. The absence of modifications results in very weak performance. Other architectures, such as ~\cite{sam2017switching}, uses VGG-16 as the density level classifier for labeling input images before sending them to the most suitable column of the MCNN, while the CP-CNN~\cite{sindagi2017generating} incorporates the result of classification with the features from density map generator. In these cases, the VGG-16 performs as an ancillary without significantly boosting the final accuracy. 
In this paper, we first remove the classification part of VGG-16 (fully-connected layers) and build the proposed CSRNet with convolutional layers in VGG-16. The output size of this front-end network is 1/8 of the original input size. If we continue to stack more convolutional layers and pooling layers (basic components in VGG-16), output size would be further shrunken, and it is hard to generate high-quality density maps. Inspired by the works~\cite{yu2015multi,chen2016deeplab,DBLP:journals/corr/ChenPSA17}, we try to deploy dilated convolutional layers as the back-end for extracting deeper information of saliency as well as maintaining the output resolution.

\subsubsection{Dilated convolution}

One of the critical components of our design is the dilated convolutional layer. A 2-D dilated convolution can be defined as follow:

\begin{equation}
y(m,n)=\sum_{i=1}^{M}\sum_{j=1}^{N}x(m+r\times i,n+r\times j)w(i,j)
\end{equation}

\begin{math}y(m,n)\end{math} is the output of dilated convolution from input \begin{math}x(m,n)\end{math} and a filter \begin{math}w(i,j)\end{math}  with the length and the width of \begin{math}M\end{math} and \begin{math}N\end{math} respectively. The parameter \begin{math}r\end{math} is the dilation rate. If $r=1$, a dilated convolution turns into a normal convolution.

\begin{figure}[t]
\begin{center}
\includegraphics[width=\linewidth]{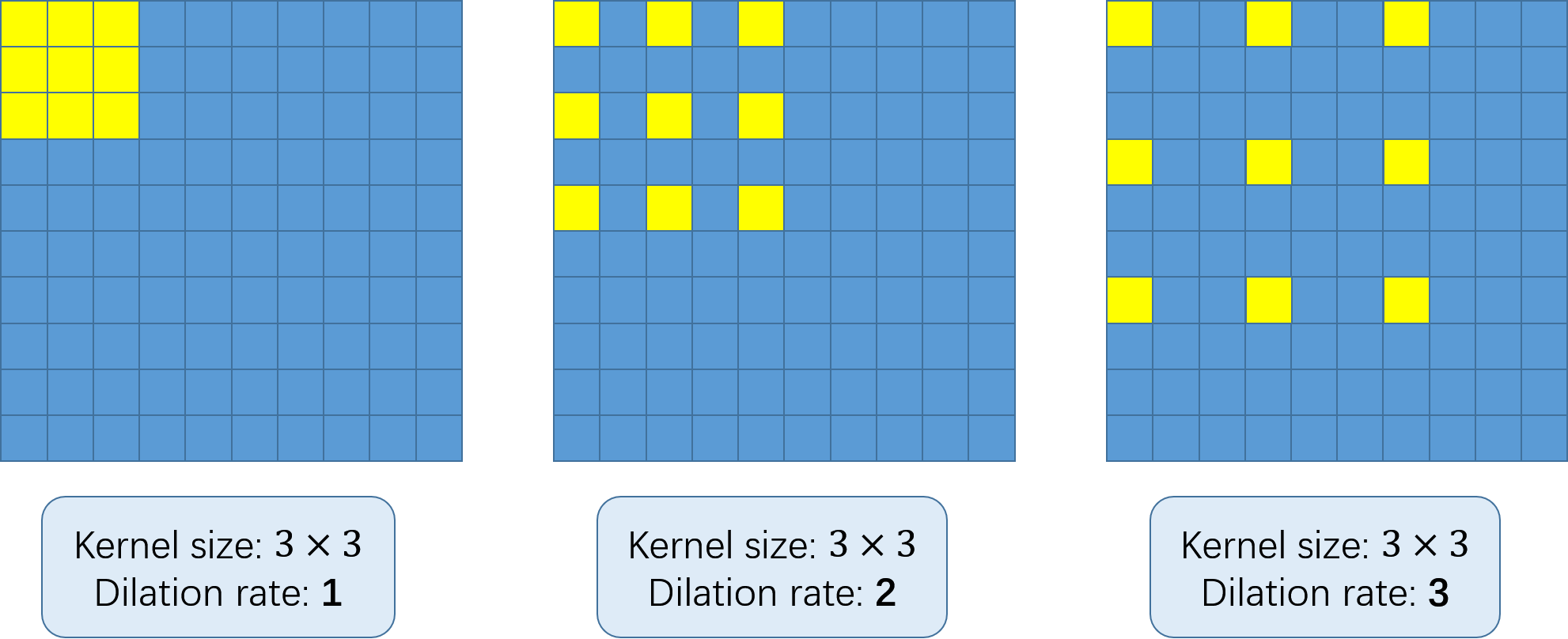}
\end{center}
\vspace{-8pt}
   \caption{$3\times3$ convolution kernels with different dilation rate as 1, 2, and 3.}
\label{fig:2}
\vspace{-6pt}
\end{figure}

Dilated convolutional layers have been demonstrated in segmentation tasks with significant improvement of accuracy~\cite{yu2015multi,chen2016deeplab,DBLP:journals/corr/ChenPSA17} and it is a good alternative of pooling layer.
Although pooling layers (e.g., max and average pooling) are widely used for maintaining invariance and controlling overfitting, they also dramatically reduce the spatial resolution meaning the spatial information of feature map is lost. Deconvolutional layers~\cite{zeiler2010deconvolutional,noh2015learning} can alleviate the loss of information, but the additional complexity and execution latency may not be suitable for all cases. Dilated convolution is a better choice, which uses sparse kernels (as shown in Fig.~\ref{fig:2}) to alternate the pooling and convolutional layer. 
This character enlarges the receptive field without increasing the number of parameters or the amount of computation (e.g., adding more convolutional layers can make larger receptive fields but introduce more operations). In dilated convolution, a small-size kernel with \begin{math}k\times k\end{math} filter is enlarged to \begin{math}k+(k-1)(r-1)\end{math} with dilated stride \begin{math}r\end{math}. Thus it allows flexible aggregation of the multi-scale contextual information while keeping the same resolution. Examples can be found in Fig.~\ref{fig:2} where normal convolution gets $3\times3$ receptive field and two dilated convolutions deliver $5\times5$ and $7\times7$ receptive fields respectively.

\begin{figure}[t]
\begin{center}
\includegraphics[width=\linewidth]{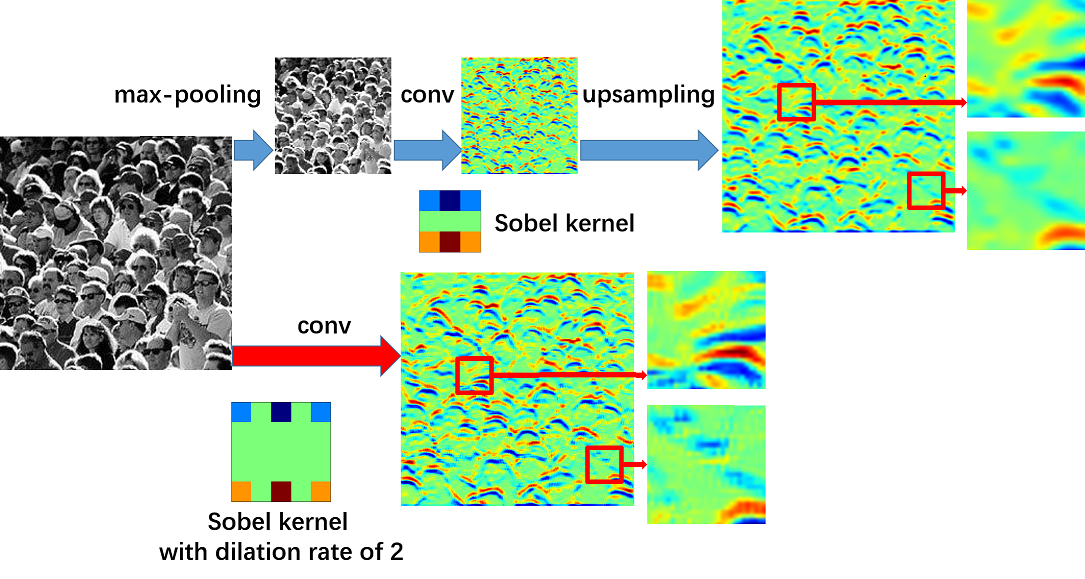}
\end{center}
\vspace{-10pt}
   \caption{Comparison between dilated convolution and max-pooling, convolution, upsampling. The $3\times3$ Sobel kernel is used in both operations while the dilation rate is 2.}
   \vspace{-10pt}
\label{fig:3}
\end{figure}

For maintaining the resolution of feature map, the dilated convolution shows distinct advantages compared to the scheme of using convolution $+$ pooling $+$ deconvolution. We pick one example for illustration in Fig.~\ref{fig:3}. The input is an image of crowds, and it is processed by two approaches separately for generating output with the same size. In the first approach, input is downsampled by a max pooling layer with factor 2, and then it is passed to a convolutional layer with a $3\times3$ Sobel kernel. Since the generated feature map is only 1/2 of the original input, it needs to be upsampled by the deconvolutional layer (bilinear interpolation). In the other approach, we try dilated convolution and adapt the same $3\times3$ Sobel kernel to a dilated kernel with a factor $=2$ stride. The output is shared the same dimension as the input (meaning pooling and deconvolutional layers are not required). Most importantly, the output from dilated convolution contains more detailed information (referring to the portions we zoom in).

\subsubsection{Network Configuration}
We propose four network configurations of CSRNet in Table~\ref{tab:3} which have the same front-end structure but different dilation rate in the back-end. 
Regarding the front-end, we adapt a VGG-16 network~\cite{simonyan2014very} (except fully-connected layers) and only use \begin{math}
3\times3\end{math} kernels. According to ~\cite{simonyan2014very}, using more convolutional layers with small kernels is more efficient than using fewer layers with larger kernels when targeting the same size of receptive field .

By removing the fully-connected layers, we try to determine the number of layers we need to use from VGG-16. The most critical part relays on the tradeoff between accuracy and the resource overhead (including training time, memory consumption, and the number of parameters). Experiment shows a best tradeoff can be achieved when keeping the first ten layers of VGG-16~\cite{simonyan2014very} with only three pooling layers instead of five to suppress the detrimental effects on output accuracy caused by the pooling operation.
Since the output (density maps) of CSRNet is smaller (1/8 of input size), we choose bilinear interpolation with the factor of 8 for scaling and make sure the output shares the same resolution as the input image. With the same size, CSRNet generated results are comparable with the ground truth results using the PSNR (Peak Signal-to-Noise Ratio) and SSIM (Structural Similarity in Image~\cite{wang2004image}). 

\subsection{Training method}
\label{subsec:Training method}
In this section, we provide specific details of CSRNet training. By taking advantage of the regular CNN network (without branch structures), CSRNet is easy to implement and fast to deploy.

\subsubsection{Ground truth generation}
\label{subsec:Ground truth}

Following the method of generating density maps in ~\cite{zhang2016single}, we use the geometry-adaptive kernels to tackle the highly congested scenes. By blurring each head annotation using a Gaussian kernel (which is normalized to 1), we generate the ground truth considering the spatial distribution of all images from each dataset. The geometry-adaptive kernel is defined as:

\begin{equation}
F(\bm{x})=\sum_{i=1}^{N}\delta(\bm{x}-\bm{x}_{i})\times G_{\sigma_{i}}(\bm{x}){,\ with\ }  \sigma_{i}=\beta\overline{d_{i}}
\end{equation}

For each targeted object \begin{math}\bm{x}_{i}\end{math} in the ground truth \begin{math}\delta\end{math}, we use \begin{math}\overline{d_{i}}\end{math} to indicate the average distance of \begin{math}k\end{math} nearest neighbors. To generate the density map, we convolve \begin{math}\delta(\bm{x}-\bm{x}_{i})\end{math} with a Gaussian kernel with parameter \begin{math}\sigma_{i}\end{math} (standard deviation), where \begin{math}\bm{x}\end{math} is the position of pixel in the image. In experiment, we follow the configuration in ~\cite{zhang2016single} where \begin{math}\beta=0.3\end{math} and \begin{math}k=3\end{math}. For input with sparse crowd, we adapt the Gaussian kernel to the average head size to blur all the annotations. The setups for different datasets are shown in Table ~\ref{tab:2}.

\begin{table}
\begin{center}
\small
\begin{tabular}{|l|c|}
\hline
Dataset & Generating method                  \\\hline\hline
ShanghaiTech Part\_A~\cite{zhang2016single} & \multirow{2}{*}{Geometry-adaptive kernels} \\ \cline{1-1}
UCF\_CC\_50~\cite{idrees2013multi} &                   \\ \hline
ShanghaiTech Part\_B~\cite{zhang2016single}  &      Fixed kernel:  $\sigma=15$\\ \hline
TRANCOS~\cite{TRANCOSdataset_IbPRIA2015}  &  Fixed kernel:  $\sigma=10$                  \\ \hline
The WorldExpo'10~\cite{zhang2015cross} & \multirow{2}{*}{Fixed kernel:  $\sigma=3$    } \\ \cline{1-1}
The UCSD~\cite{4587569} &                   \\ \hline

\end{tabular}
\end{center}
\vspace{-8pt}
\caption{The ground truth generating methods for different datasets}
\label{tab:2}
\vspace{-10pt}
\end{table}

\subsubsection{Data augmentation}
We crop 9 patches from each image at different locations with 1/4 size of the original image. The first four patches contain four quarters of the image without overlapping while the other five patches are randomly cropped from the input image. After that, we mirror the patches so that we double the training set.

\subsubsection{Training details}

We use a straightforward way to train the CSRNet as an end-to-end structure. The first 10 convolutional layers are fine-tuned from a well-trained VGG-16~\cite{simonyan2014very}. For the other layers, the initial values come from a Gaussian initialization with 0.01 standard deviation. Stochastic gradient descent (SGD) is applied with fixed learning rate at 1e-6 during training. Also, we choose the Euclidean distance to measure the difference between the ground truth and the  estimated density map we generated which is similar to other works~\cite{boominathan2016crowdnet,zhang2016single,sam2017switching}. The loss function is given as follow:

\begin{equation}
L(\Theta)=\frac{1}{2N}\sum_{i=1}^{N}{\parallel~Z(X_i;\Theta)-Z_{i}^{GT}\parallel}_2^2
\end{equation}

 where \begin{math}N\end{math} is the size of training batch and \begin{math}Z(X_i;\Theta)\end{math} is the output generated by CSRNet with parameters shown as \begin{math}\Theta\end{math}. \begin{math}X_i\end{math} represents the input image while \begin{math}Z_{i}^{GT}\end{math} is the ground truth result of the input image \begin{math}X_i\end{math}.

\begin{table}
\small
\begin{center}

\begin{tabular}{|c|c|c|c|}
\hline
\multicolumn{4}{|c|}{Configurations of CSRNet}                                                                                                                                                                                                                                                                                                                                                                                                                                                                  \\ \hline
A                                                                                                                        & B                                                                                                                        & C                                                                                                                        & D                                                                                                                        \\ \hline\hline
\multicolumn{4}{|c|}{input(unfixed-resolution color image)}                                                                                                                                                                                                                                                                                                                                                                                                                                               \\ \hline
\multicolumn{4}{|c|}{\begin{tabular}[c]{@{}c@{}}front-end\\ (fine-tuned from VGG-16)\end{tabular}}                                                                                                                                                                                                                                                                                                                                                                                                        \\ \hline
\multicolumn{4}{|c|}{\begin{tabular}[c]{@{}c@{}}conv3-64-1\\ conv3-64-1\end{tabular}}                                                                                                                                                                                                                                                                                                                                                                                                                     \\ \hline
\multicolumn{4}{|c|}{max-pooling}                                                                                                                                                                                                                                                                                                                                                                                                                                                                         \\ \hline
\multicolumn{4}{|c|}{\begin{tabular}[c]{@{}c@{}}conv3-128-1\\ conv3-128-1\end{tabular}}                                                                                                                                                                                                                                                                                                                                                                                                                   \\ \hline
\multicolumn{4}{|c|}{max-pooling}                                                                                                                                                                                                                                                                                                                                                                                                                                                                         \\ \hline
\multicolumn{4}{|c|}{\begin{tabular}[c]{@{}c@{}}conv3-256-1\\ conv3-256-1\\ conv3-256-1\end{tabular}}                                                                                                                                                                                                                                                                                                                                                                                                     \\ \hline
\multicolumn{4}{|c|}{max-pooling}                                                                                                                                                                                                                                                                                                                                                                                                                                                                         \\ \hline
\multicolumn{4}{|c|}{\begin{tabular}[c]{@{}c@{}}conv3-512-1\\ conv3-512-1\\ conv3-512-1\end{tabular}}                                                                                                                                                                                                                                                                                                                                                                                                     \\ \hline
\multicolumn{4}{|c|}{back-end (four different configurations)}                                                                                                                                                                                                                                                                                                                                                                                                                                                                            \\ \hline
\begin{tabular}[c]{@{}c@{}}conv3-512-1\\ conv3-512-1\\ conv3-512-1\\ conv3-256-1\\ conv3-128-1\\ conv3-64-1\end{tabular} & \begin{tabular}[c]{@{}c@{}}conv3-512-2\\ conv3-512-2\\ conv3-512-2\\ conv3-256-2\\ conv3-128-2\\ conv3-64-2\end{tabular} & \begin{tabular}[c]{@{}c@{}}conv3-512-2\\ conv3-512-2\\ conv3-512-2\\ conv3-256-4\\ conv3-128-4\\ conv3-64-4\end{tabular} & \begin{tabular}[c]{@{}c@{}}conv3-512-4\\ conv3-512-4\\ conv3-512-4\\ conv3-256-4\\ conv3-128-4\\ conv3-64-4\end{tabular} \\ \hline
\multicolumn{4}{|c|}{conv1-1-1}                                                                                                                                                                                                                                                                                                                                                                                                                                                                           \\ \hline
\end{tabular}
\end{center}
\caption{Configuration of CSRNet. All convolutional layers use padding to maintain the previous size. The convolutional layers' parameters are denoted as ``conv-(kernel size)-(number of filters)-(dilation rate)", max-pooling layers are conducted over a $2\times2$ pixel window with stride 2.}
\label{tab:3}
\vspace{-10pt}
\end{table}

\section{Experiments}
\label{sec:Experiments}
We demonstrate our approach in five different public datasets~\cite{zhang2016single,zhang2015cross,idrees2013multi,4587569,TRANCOSdataset_IbPRIA2015}. Compared to the previous state-of-the-art methods~\cite{sam2017switching,sindagi2017generating}, our model is smaller, more accurate, and easier to train and deploy. In this section, the evaluation metrics are introduced, and then an ablation study of ShanghaiTech Part\_A dataset is conducted to analyze the configuration of our model (shown in Table \ref{tab:3}). Along with the ablation study, we evaluate and compare our proposed method to the previous state-of-the-art methods in all these five datasets. 
The implementation of our model is based on the Caffe framework~\cite{jia2014caffe}.

\subsection{Evaluation metrics}
\label{subsec:Evaluation metrics}

The MAE and the MSE are used for evaluation which are defined as:
\begin{equation}
MAE=\frac{1}{N}\sum_{i=1}^{N}\mid C_i-C_i^{GT} \mid
\end{equation}
\begin{equation}
MSE=\sqrt{\frac{1}{N}\sum_{i=1}^{N}\mid C_i-C_i^{GT} \mid^2}
\end{equation}

where \begin{math}N\end{math} is the number of images in one test sequence and \begin{math}C_{i}^{GT}\end{math}  is the ground truth of counting. \begin{math}C_{i}\end{math} represents the estimated count which is defined as follows:
\begin{equation}
C_i=\sum_{l=1}^{L}\sum_{w=1}^{W}z_{l,w}
\end{equation}

\begin{math}L\end{math} and \begin{math}W\end{math} show the length and width of the density map respectively while \begin{math}z_{l,w}\end{math} is the pixel at \begin{math}(l,w)\end{math} of the generated density map. \begin{math}C_i\end{math} means the estimated counting number for image \begin{math}X_i\end{math}.

We also use the PSNR and SSIM to evaluate the quality of the output density map on ShanghaiTech Part\_A dataset. To calculate the PSNR and SSIM, we follow the preprocess given by~\cite{sindagi2017generating}, which includes the density map resizing (same size with the original input) with interpolation and normalization for both ground truth and predicted density map. 

\subsection{Ablations on ShanghaiTech Part\_A}
\label{subsec:Ablation}

In this subsection, we perform an ablation study to analyze the four configurations of the CSRNet on ShanghaiTech Part\_A dataset~\cite{zhang2016single} which is a new large-scale crowd counting dataset including 482 images for congested scenes with 241,667 annotated persons. It is challenging to count from these images because of the extremely congested scenes, the varied perspective, and the unfixed resolution. %
These four configurations are shown in Table~\ref{tab:3}. CSRNet A is the network with all the dilation rate of 1. CSRNet B and D maintain the dilation rate of 2 and 4 in their back-end respectively while CSRNet C combines the dilated rate of 2 and 4. The number of parameters of these four models are the same as 16.26M. We intend to compare the results by using different dilation rates.
After training on Shanghai Part\_A dataset using the method mentioned in Sec.~\ref{subsec:Training method}, we perform the evaluation metrics defined in Sec.~\ref{subsec:Evaluation metrics}. We try  dropout~\cite{JMLR:v15:srivastava14a} for preventing the potential overfitting problem but there is no significant improvement. So we do not include dropout in our model. The detailed evaluation results are shown in Table~\ref{tab:4}, where CSRNet B achieves the lowest error (the highest accuracy). Therefore, we use CSRNet B as the proposed CSRNet for the following experiments.

\begin{table}[!]
\begin{center}
\begin{tabular}{|l|c|c|}
\hline
Architecture & MAE & MSE  \\
\hline\hline
CSRNet A &69.7 &116.0
  \\
\hline
CSRNet B &\textbf{68.2} &\textbf{115.0}  \\
\hline
CSRNet C& 71.91&120.58  \\
\hline
CSRNet D& 75.81& 120.82 \\
\hline

\end{tabular}
\vspace{-6pt}
\end{center}
\caption{Comparison of architectures on ShanghaiTech Part\_A dataset}
\label{tab:4}
\vspace{-10pt}
\end{table}

\subsection{Evaluation and comparison}

\subsubsection{ShanghaiTech dataset}

\begin{figure}[!t]
\begin{center}
\includegraphics[width=\linewidth]{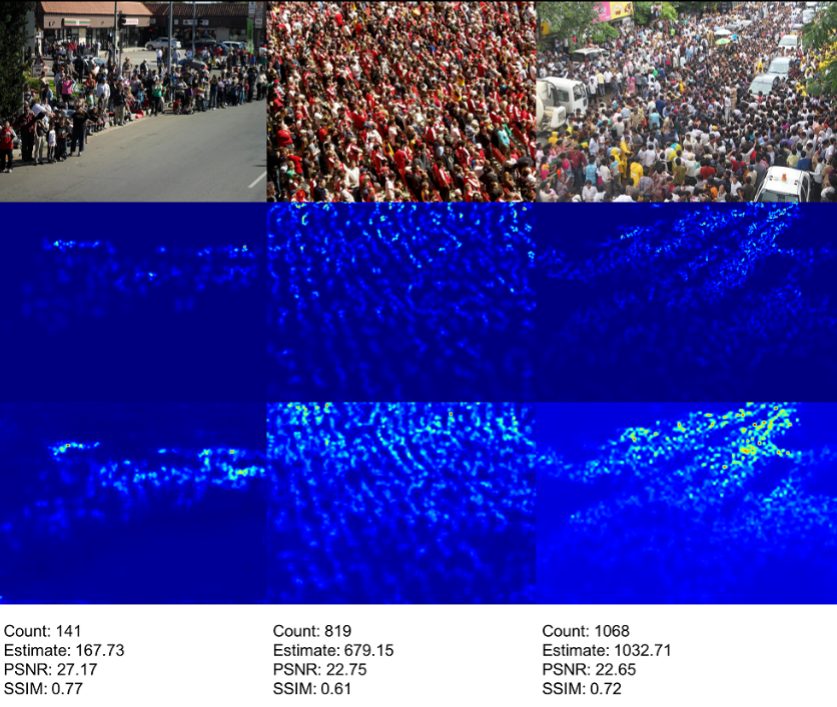}
\end{center}
   \caption{The first row shows the samples of the testing set in ShanghaiTech Part\_A dataset. The second row shows the ground truth for each sample while the third row presents the generated density map by CSRNet.}
   \label{fig:parta}
\end{figure}

ShanghaiTech crowd counting dataset contains 1198 annotated images with a total amount of 330,165 persons~\cite{zhang2015cross}. This dataset consists of two parts as Part\_A containing 482 images with highly congested scenes randomly downloaded from the Internet while Part\_B includes 716 images with relatively sparse crowd scenes taken from streets in Shanghai. Our method is evaluated and compared to other six recent works and results are shown in Table~\ref{tab:5}.
It indicates that our method achieves the lowest MAE (the highest accuracy) in Part\_A compared to other methods and we get 7\% lower MAE than the state-of-the-art solution called CP-CNN. CSRNet also delivers 47.3\% lower MAE in Part\_B compared to the CP-CNN. To evaluate the quality of generated density map, we compare our method to the MCNN and the CP-CNN using Part\_A dataset and we follow the evaluation metrics in Sec.~\ref{subsec:Training method}. Samples of the test cases can be found in Fig~\ref{fig:parta}. Results are shown in Table~\ref{tab:6} which indicates CSRNet achieves the highest SSIM and PSNR. We also report the quality result of ShanghaiTech dataset in Table~\ref{tab:10}.

\begin{table}[!t]
\begin{center}
\small
\begin{tabular}{|l|c|c|c|c|}
\hline
 & \multicolumn{2}{c|}{Part\_A} & \multicolumn{2}{c|}{Part\_B} \\ \hline
Method &    MAE       &MSE&MAE&MSE\\ \hline\hline
Zhang \etal.~\cite{zhang2015cross} &181.8&277.7&32.0&49.8        \\ \hline
Marsden \etal.~\cite{marsden2016fully} &126.5&173.5&23.8&33.1\\ \hline
MCNN~\cite{zhang2016single} &110.2&173.2&26.4&41.3\\ \hline
Cascaded-MTL~\cite{sindagi2017cnn} &101.3&152.4&20.0&31.1\\ \hline
Switching-CNN~\cite{sam2017switching} &90.4&135.0&21.6&33.4\\ \hline
CP-CNN~\cite{sindagi2017generating} &73.6&\textbf{106.4}&20.1&30.1\\ \hline
CSRNet (ours) &\textbf{68.2}&115.0&\textbf{10.6}&\textbf{16.0}\\ \hline
\end{tabular}
\end{center}
\vspace{-4pt}
\caption{Estimation errors on ShanghaiTech dataset}
\label{tab:5}
\end{table}

\begin{table}[!]
\begin{center}
\small
\begin{tabular}{|l|c|c|}
\hline
Method &PSNR  &SSIM  \\ \hline\hline
 MCNN~\cite{zhang2016single}& 21.4 & 0.52 \\ \hline
 CP-CNN~\cite{sindagi2017generating}& 21.72 &0.72  \\ \hline
 CSRNet (ours)&\textbf{23.79}&  \textbf{0.76}\\ \hline
\end{tabular}

\end{center}
\caption{Quality of density map on ShanghaiTech Part\_A dataset}
\label{tab:6}
\end{table}

\subsubsection{UCF\_CC\_50 dataset}
UCF\_CC\_50 dataset includes 50 images with different perspective and resolutions~\cite{idrees2013multi}. The number of annotated persons per image ranges from 94 to 4543 with an average number of 1280. 5-fold cross-validation is performed following the standard setting in~\cite{idrees2013multi}. Result comparisons of MAE and MSE are listed in Table~\ref{tab:7} while the quality of generated density map can be found in Table~\ref{tab:10}.

\begin{table}[!t]
\begin{center}
\small
\begin{tabular}{|l|c|c|}
\hline
Method & MAE & MSE \\ \hline\hline
Idrees \etal.~\cite{idrees2013multi} & 419.5 & 541.6 \\ \hline
 Zhang \etal.~\cite{zhang2015cross}& 467.0 &498.5  \\ \hline
 MCNN~\cite{zhang2016single}&  377.6&509.1  \\ \hline
 Onoro \etal.~\cite{onoro2016towards} Hydra-2s& 333.7 & 425.2 \\ \hline
 Onoro \etal.~\cite{onoro2016towards} Hydra-3s& 465.7 &371.8  \\ \hline
 Walach \etal.~\cite{walach2016learning}&364.4  &341.4  \\ \hline
 Marsden \etal.~\cite{marsden2016fully}& 338.6 &  424.5\\ \hline
 Cascaded-MTL~\cite{sindagi2017cnn}& 322.8 & 397.9 \\ \hline
 Switching-CNN~\cite{sam2017switching}& 318.1 & 439.2 \\ \hline
 CP-CNN~\cite{sindagi2017generating}& 295.8 & \textbf{320.9} \\ \hline
 CSRNet (ours)&\textbf{266.1}  & 397.5 \\ \hline
\end{tabular}
\end{center}

\caption{Estimation errors on UCF\_CC\_50 dataset}

\label{tab:7}
\end{table}

\subsubsection{The WorldExpo'10 dataset}
The WorldExpo'10 dataset~\cite{zhang2015cross} consists of 3980 annotated frames from 1132 video sequences captured by 108 different surveillance cameras. This dataset is divided into a training set (3380 frames) and a testing set (600 frames) from five different scenes. The region of interest (ROI) is provided for the whole dataset. Each frame and its dot maps are masked with ROI during preprocessing, and we train our model following the instructions given in Sec.~\ref{subsec:Training method}. Results are shown in Table~\ref{tab:8}. The proposed CSRNet delivers the best accuracy in 4 out of 5 scenes and it achieves the best accuracy on average. 

\begin{table}[!h]
\scriptsize
\begin{center}
\begin{tabular}{|l|c|c|c|c|c|c|}
\hline
 Method&Sce.1  &  Sce.2&Sce.3  &Sce.4  &Sce.5  &Avg.  \\ \hline\hline
Chen \etal.~\cite{chen2013cumulative} &2.1  & 55.9 &  9.6&11.3  & 3.4 &16.5  \\ \hline
 Zhang \etal.~\cite{zhang2015cross}&9.8  &14.1  &14.3  &22.2  & 3.7 & 12.9 \\ \hline
 MCNN~\cite{zhang2016single}&3.4  &20.6  &12.9  &13.0  &8.1  & 11.6 \\ \hline
 Shang \etal.~\cite{shang2016end}&7.8  & 15.4 &14.9  &11.8  & 5.8 & 11.7 \\ \hline
 Switching-CNN~\cite{sam2017switching}&4.4  &15.7  &10.0  &11.0  &5.9  &9.4  \\ \hline
 CP-CNN~\cite{sindagi2017generating}&2.9  &14.7  & 10.5 &\textbf{10.4}  & 5.8 &8.86  \\ \hline
 CSRNet (ours)&\textbf{2.9}  &\textbf{11.5}  &\textbf{8.6}  &16.6  &\textbf{3.4}  &\textbf{8.6}  \\ \hline
\end{tabular}
\end{center}
\caption{Estimated errors on the WorldExpo'10 dataset}
\label{tab:8}
\vspace{-10pt}
\end{table}

\subsubsection{The UCSD dataset}
The UCSD dataset~\cite{4587569} has 2000 frames captured by surveillance cameras. These scenes contain sparse crowd varying from 11 to 46 persons per image. The region of interest (ROI) is also provided. Because the resolution of each frame is fixed and small ($238\times 158$), it is difficult to generate a high-quality density map after frequent pooling operations. So we preprocess the frames by using bilinear interpolation to resize them into $952\times 632$. Among the 2000 frames, we use frames 601 through 1400 as training set and the rest of them as testing set according to ~\cite{4587569}. Before blurring the annotation as we mentioned in Sec.~\ref{subsec:Training method}, all the frames and the corresponding dot maps are masked with ROI. The accuracy of running UCSD dataset is shown in Table~\ref{tab:111} and we outperform most of the previous methods except MCNN in the MAE category.
Results indicate that our method can perform not only counting tasks for extremely dense crowds but also tasks for relative sparse scenes. Also, we provides the quality of generated density map in Table~\ref{tab:10}.

\begin{table}[!t]
\begin{center}
\small
\begin{tabular}{|l|c|c|}
\hline
Method & MAE & MSE \\ \hline\hline
 Zhang \etal.~\cite{zhang2015cross}& 1.60 &3.31  \\ \hline
 CCNN~\cite{onoro2016towards} CCNN& 1.51 &-  \\ \hline
 Switching-CNN~\cite{sam2017switching}& 1.62 & 2.10 \\ \hline
 FCN-rLSTM~\cite{DBLP:journals/corr/ZhangWCM17aa}& 1.54 & 3.02 \\ \hline

 CSRNet (ours)&1.16 & 1.47 \\ \hline
 MCNN~\cite{zhang2016single}& \textbf{1.07}&\textbf{1.35}  \\ \hline

\end{tabular}
\end{center}
\caption{ Estimation errors on the UCSD dataset}
\label{tab:111}
\vspace{-10pt}
\end{table}

\subsubsection{TRANCOS dataset}
Beyond the crowd counting, we setup an experiment on the TRANCOS dataset~\cite{TRANCOSdataset_IbPRIA2015} for vehicle counting to demonstrate the robustness and generalization of our approach. TRANCOS is a public traffic dataset containing 1244 images of different congested traffic scenes captured by surveillance cameras with 46796 annotated vehicles. Also, the region of interest (ROI) is provided for the evaluation. The perspectives of images are not fixed and the images are collected from very different scenarios. 
The Grid Average Mean Absolute Error (GAME)~\cite{TRANCOSdataset_IbPRIA2015} is used for evaluation in this test. The GAME is defined as follow:

\begin{equation}
GAME(L)=\frac{1}{N}\sum_{n=1}^{N}(\sum_{l=1}^{^{4^{L}}}\left | D_{I_{n}}^{l} -D_{I_{n}^{gt}}^{l}\right |)
\end{equation}

where \begin{math} 
N
\end{math} is the number of images in testing set, and \begin{math}
 D_{I_{n}}^{l}
\end{math} is the estimated result of the input image \begin{math}
n
\end{math} within region \begin{math}
l
\end{math}. \begin{math}
D_{I_{n}^{gt}}^{l}
\end{math} is the corresponding ground truth result. For a specific level \begin{math}
L
\end{math}, the GAME(\begin{math}
L
\end{math}) subdivides the image using a grid of \begin{math}
4^{L}
\end{math} non-overlapping regions which cover the full image, and the error is computed as the sum of the MAE in each of these regions. When \begin{math}
L=0
\end{math}, the GAME is equivalent to the MAE metric.

\begin{figure}[!t]
\begin{center}
\includegraphics[width=\linewidth]{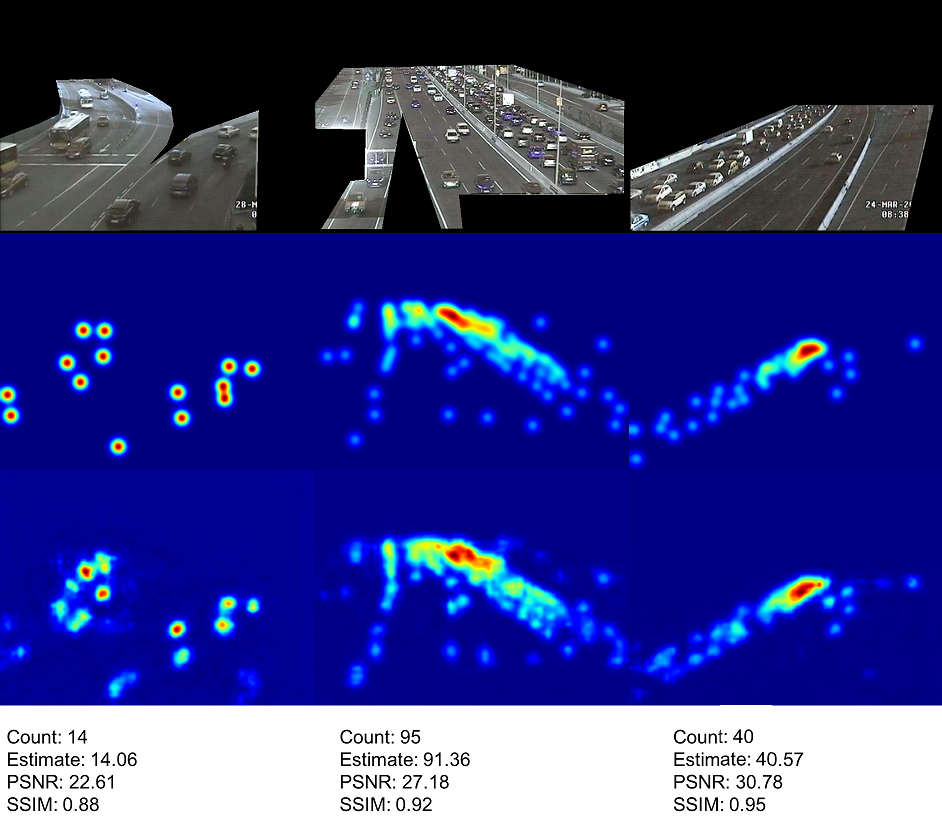}
\end{center}
   \caption{The first row shows samples of the testing set in TRANCOS~\cite{TRANCOSdataset_IbPRIA2015} dataset with ROI. The second row shows the ground truth for each sample. The third row shows the generated density map by CSRNet.}
\vspace{-10pt} 
\label{fig:trancos}

\end{figure}

We compare our approach with the previous state-of-the-art methods~\cite{6460719,lempitsky2010learning,onoro2016towards,DBLP:journals/corr/ZhangWCM17aa}. The method in~\cite{onoro2016towards} uses the MCNN-like network to generate the density map while the model in~\cite{DBLP:journals/corr/ZhangWCM17aa} deploys a combination of fully convolutional neural networks (FCN) and a long short-term memory network (LSTM). Results are shown in Table~\ref{tab:9} with three examples shown in Fig.~\ref{fig:trancos}. Our model achieves a significant improvement on four different GAME metrics. Compared to the result from ~\cite{onoro2016towards}, CSRNet delivers 67.7\% lower GAME(0), 60.1\% lower GAME(1), 48.7\% lower GAME(2), and 22.2\% lower GAME(3), which is the best solution. We also present the quality of generated density map in Table~\ref{tab:10}.

\begin{table}[!h]
\scriptsize
\begin{center}
\begin{tabular}{|l|c|c|c|c|}
\hline
 Method& GAME 0  &GAME 1 &GAME 2 &GAME 3 \\ \hline\hline
Fiaschi \etal.~\cite{6460719} & 17.77 & 20.14 & 23.65 &25.99  \\ \hline
Lempitsky \etal.~\cite{lempitsky2010learning} & 13.76 & 16.72 & 20.72 & 24.36 \\ \hline
 Hydra-3s~\cite{onoro2016towards} &10.99  &13.75  &16.69  &19.32  \\ \hline
  FCN-HA~\cite{DBLP:journals/corr/ZhangWCM17aa}&4.21  &-  &-  &-  \\ \hline
 CSRNet (Ours)&\textbf{3.56}  & \textbf{5.49}  & \textbf{8.57} &  \textbf{15.04}\\ \hline
\end{tabular}
\end{center}
\caption{GAME on the TRANCOS dataset}
\label{tab:9}
\vspace{-5pt}
\end{table}

\begin{table}[!h]
\begin{center}
\small
\begin{tabular}{|l|c|c|}
\hline
Dataset & PSNR &SSIM  \\ \hline\hline
ShanghaiTech Part\_A~\cite{zhang2016single} &23.79  &0.76  \\ \hline
ShanghaiTech Part\_B~\cite{zhang2016single}&27.02  &0.89  \\ \hline
UCF\_CC\_50~\cite{idrees2013multi}&18.76  & 0.52 \\ \hline
The WorldExpo'10~\cite{zhang2015cross}&26.94  & 0.92 \\ \hline
The UCSD~\cite{4587569}&20.02 & 0.86\\ \hline
TRANCOS~\cite{TRANCOSdataset_IbPRIA2015}&27.10  & 0.93 \\ \hline

\end{tabular}
\end{center}
\caption{The quality of density maps generated by CSRNet in 5 datasets}
\label{tab:10}
\vspace{-10pt}
\end{table}




\section{Conclusion}
\label{sec:Conclusion}
In this paper, we proposed a novel architecture called CSRNet for crowd counting and high-quality density map generation with an easy-trained end-to-end approach. We used the dilated convolutional layers to aggregate the multi-scale contextual information  in the congested scenes. By taking advantage of the dilated convolutional layers, CSRNet can expand the receptive field without losing resolution. We demonstrated our model in four crowd counting datasets with the state-of-the-art performance. 
We also extended our model to vehicle counting task and our model achieved the best accuracy as well.

\section{Acknowledgement}

This work was supported by the IBM-Illinois Center for Cognitive Computing System Research (C3SR) - a research collaboration as part of the IBM AI Horizons Network.

{\small
\bibliographystyle{unsrt}
\bibliography{egpaper_final}

}

\onecolumn
\section{\vspace{-4pt}Appendix: supplementary material}
\begin{figure}[b]
\begin{center}
\includegraphics[width=\linewidth]{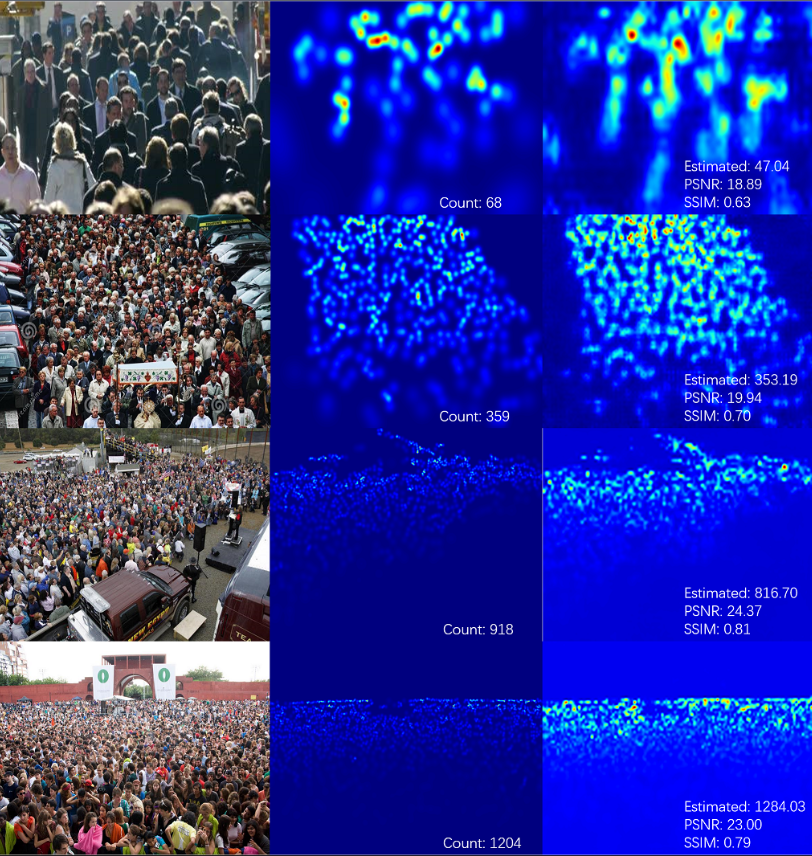}
\end{center}
  \caption{Samples generated by CSRNet from ShanghaiTech Part\_A~\cite{zhang2016single} dataset. The left column shows the original images; the medium column displays the ground truth density maps while the right column indicates our generated density maps.}
\label{appfig:1}
\vspace{-10pt}
\end{figure}

\begin{figure*}[]
\begin{center}
\includegraphics[width=\linewidth]{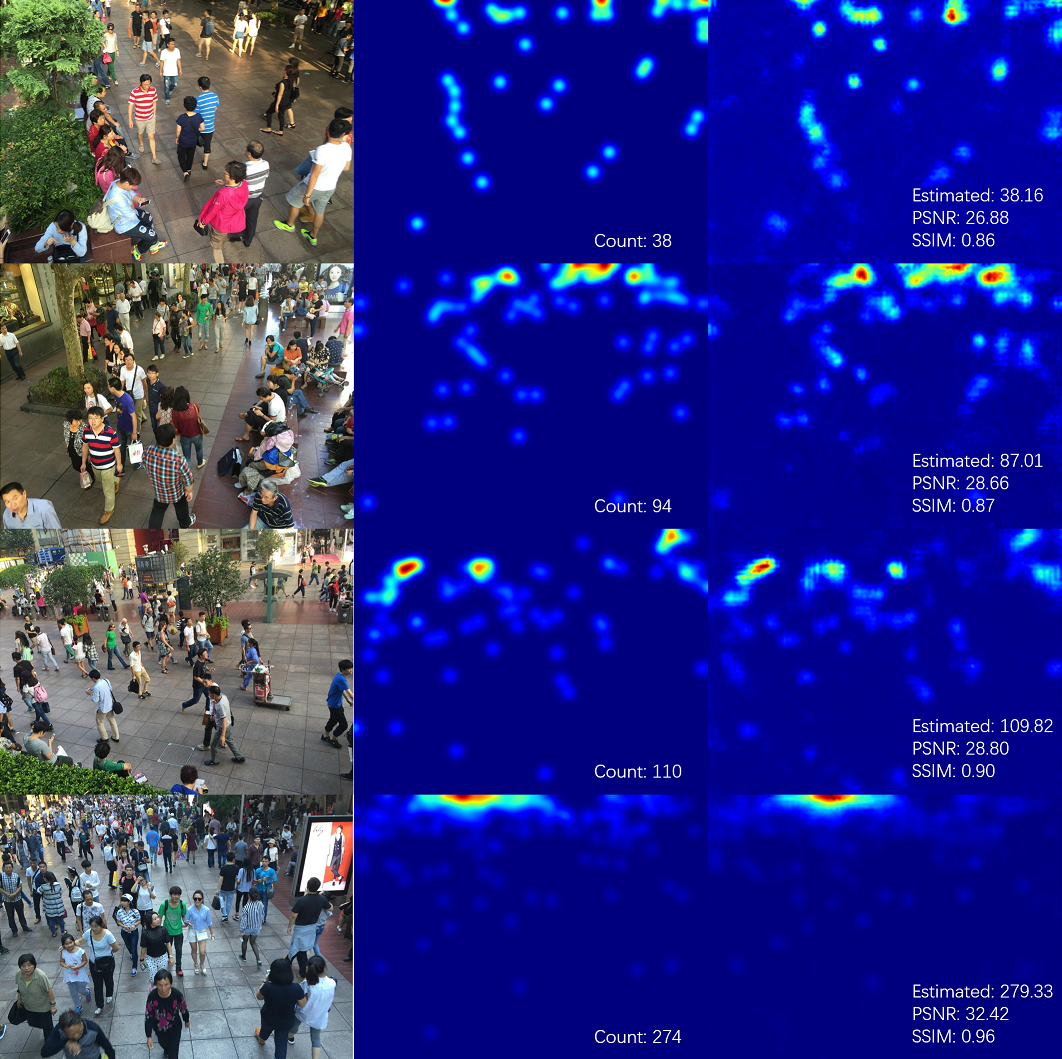}
\end{center}
  \caption{Samples generated by CSRNet from ShanghaiTech Part\_B~\cite{zhang2016single} dataset. The left column shows the original images; the medium column displays the ground truth density maps while the right column indicates our generated density maps.}
\label{appfig:2}
\vspace{-10pt}
\end{figure*}

\begin{figure*}[]
\begin{center}
\includegraphics[width=\linewidth]{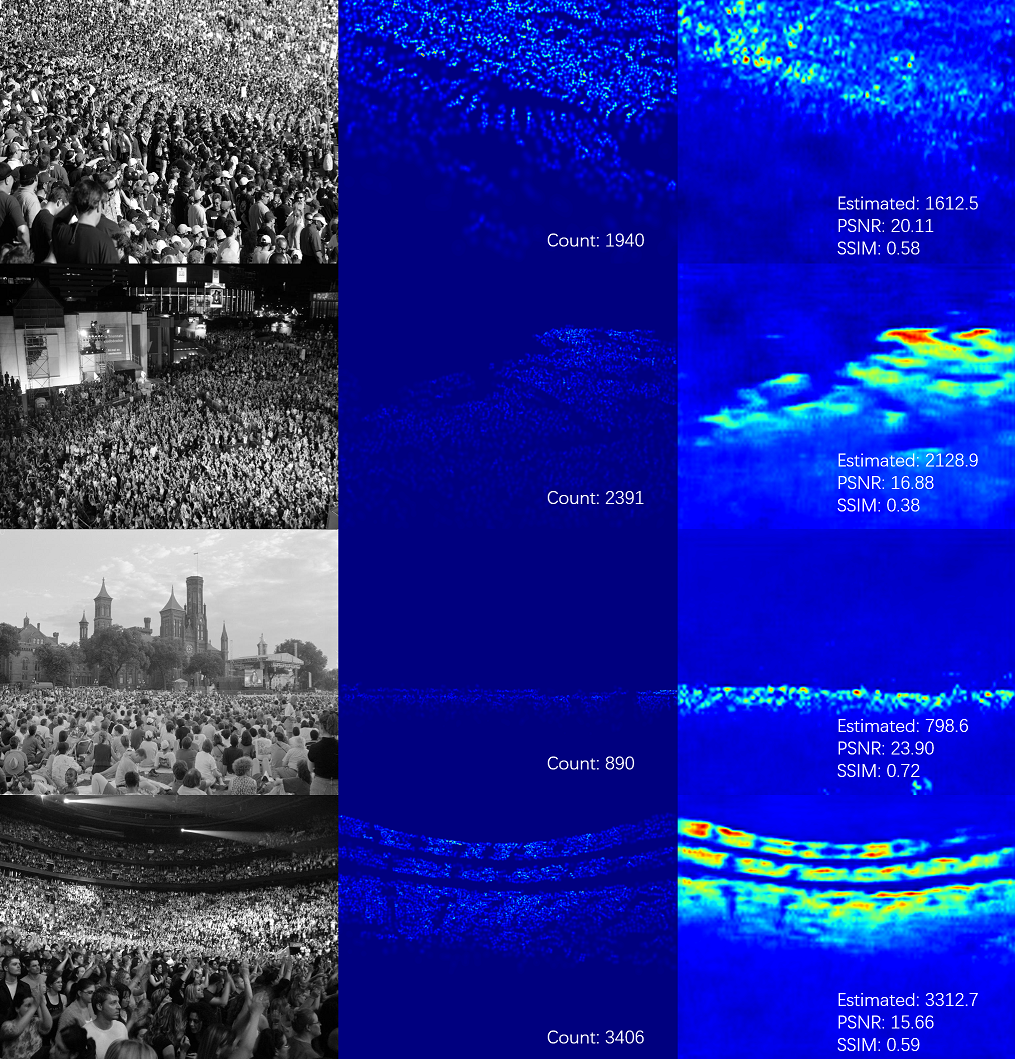}
\end{center}
  \caption{Samples generated by CSRNet from UCF\_CC\_50~\cite{idrees2013multi} dataset. The left column shows the original images; the medium column displays the ground truth density maps while the right column indicates our generated density maps.}
\label{appfig:3}
\vspace{-10pt}
\end{figure*}

\begin{figure*}[]
\begin{center}
\includegraphics[width=\linewidth]{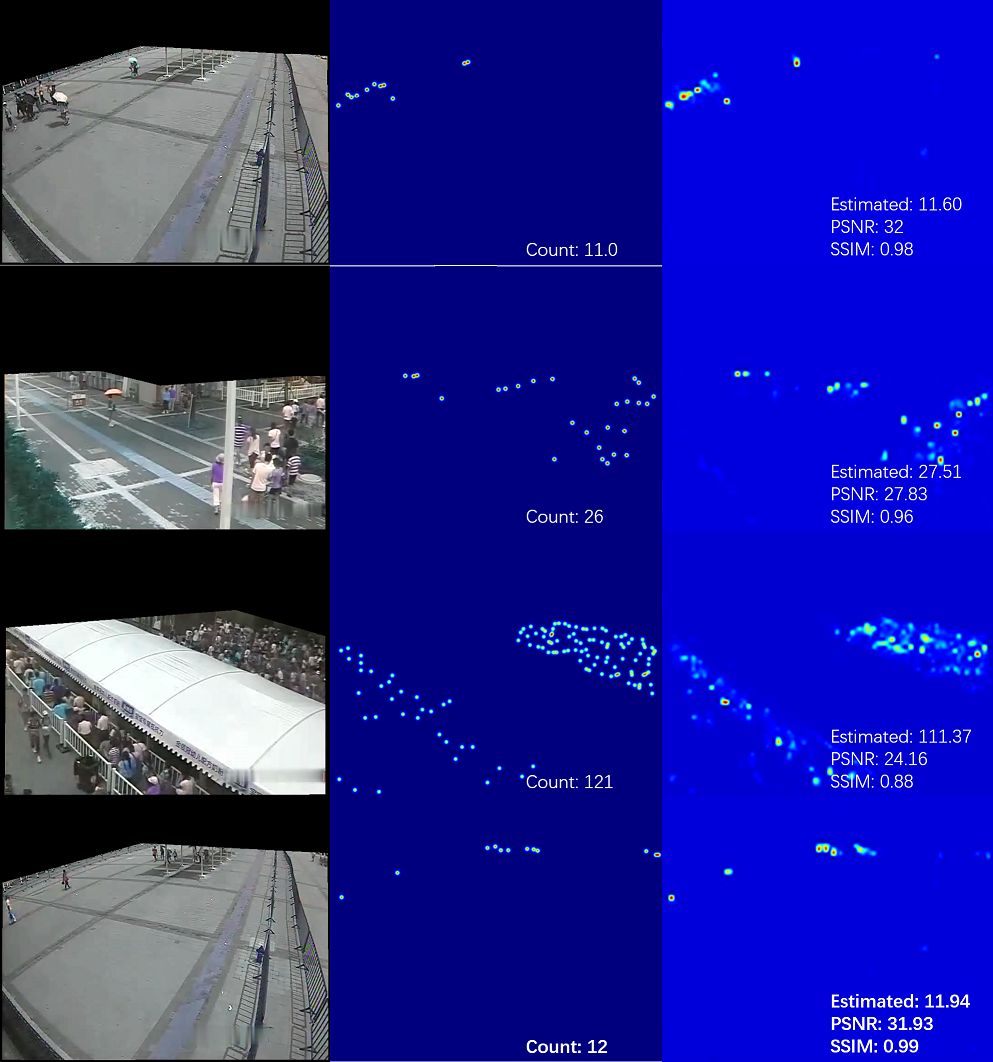}
\end{center}
  \caption{Samples generated by CSRNet from WorldExpo'10~\cite{zhang2015cross} dataset. The left column shows the images masked by the ROI (region of interest); the medium column displays the ground truth density maps and the right column indicates our generated density maps.}
\label{appfig:4}
\vspace{-10pt}
\end{figure*}

\begin{figure*}[]
\begin{center}
\includegraphics[width=\linewidth]{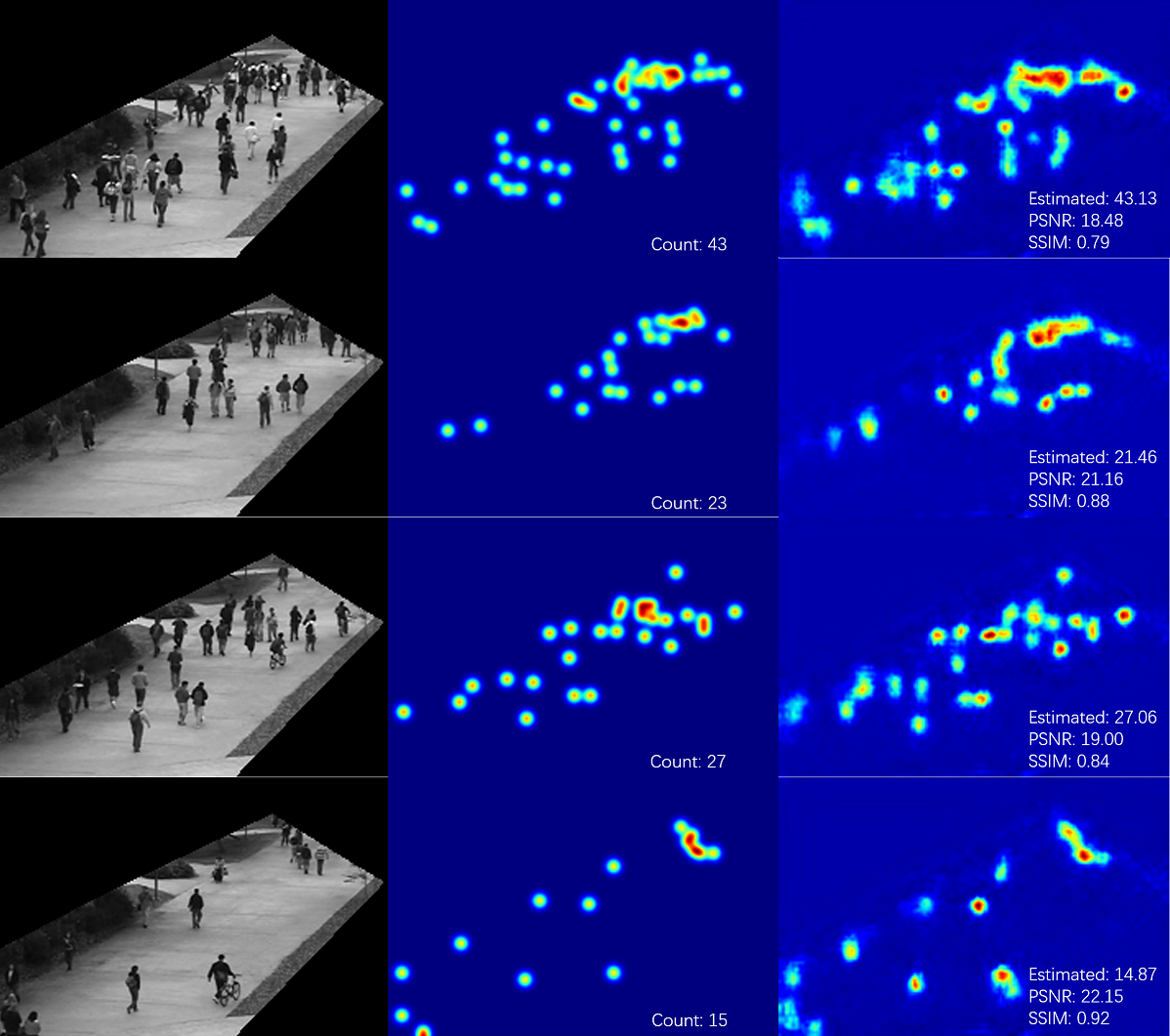}
\end{center}
  \caption{Samples generated by CSRNet from UCSD~\cite{4587569} dataset. The left column shows the images masked by the ROI; the medium column displays the ground truth density maps and the right column indicates our generated density maps.}
\label{appfig:5}
\vspace{-10pt}
\end{figure*}

\begin{figure*}[]
\begin{center}
\includegraphics[width=\linewidth]{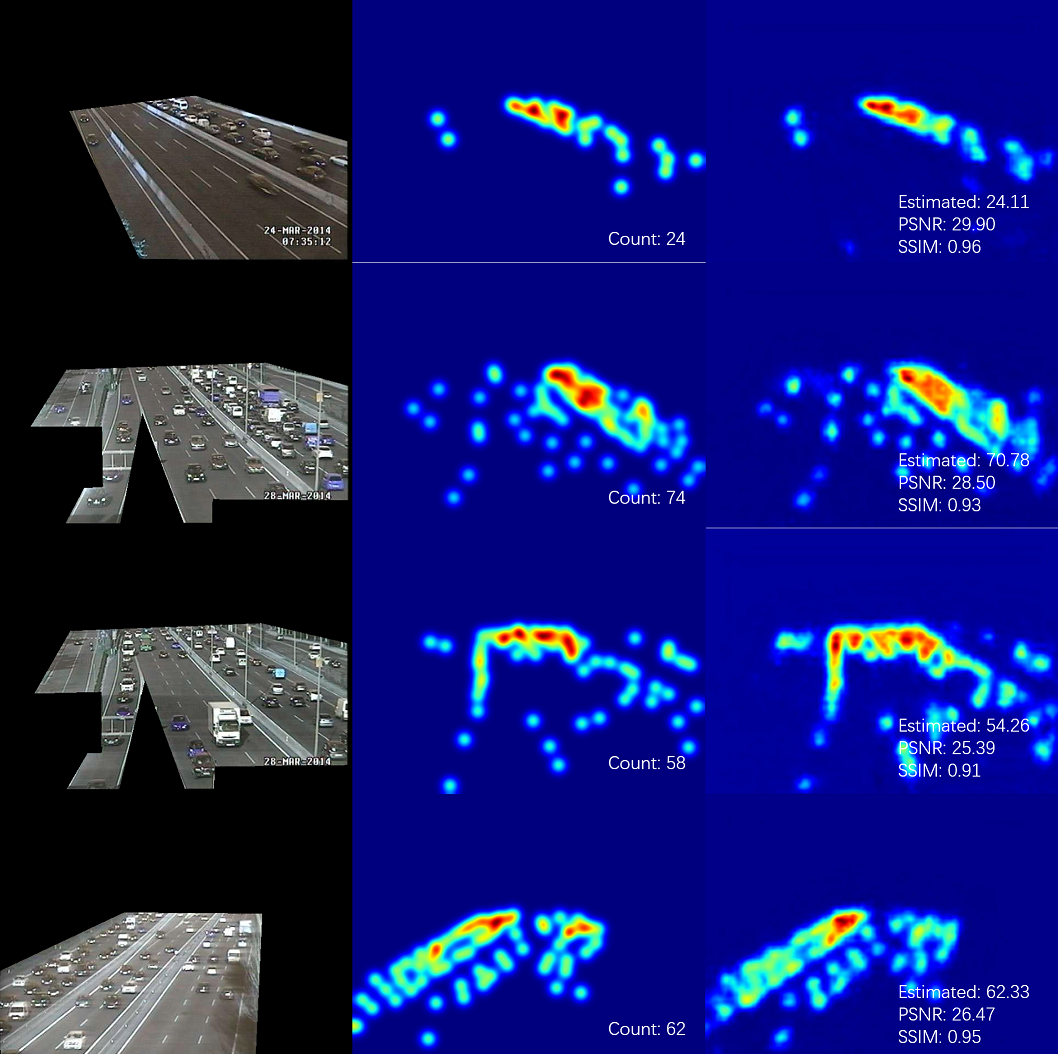}
\end{center}
  \caption{Samples generated by CSRNet from TRANCOS~\cite{TRANCOSdataset_IbPRIA2015} dataset. The left column shows the images masked by the ROI; the medium column displays the ground truth density maps while the right column shows the generated density maps.}
\label{appfig:6}
\vspace{-10pt}
\end{figure*}

In this appendix, additional results generated by CSRNet from five datasets (ShanghaiTech~\cite{zhang2016single}, UCF\_CC\_50~\cite{idrees2013multi}, WorldExpo'10~\cite{zhang2015cross}, UCSD~\cite{4587569}, and TRANCOS~\cite{TRANCOSdataset_IbPRIA2015}) are presented to demonstrate the validity of our design. Two criteria are used as the PSNR (Peak Signal-to-Noise Ratio) and the SSIM (Structural Similarity in Image~\cite{wang2004image} to evaluate our design's quality of generated density maps. Samples from these 5 datasets are shown in Fig.~\ref{appfig:1} to Fig.~\ref{appfig:6}, which represent a variety of density levels.
\vspace{-6pt}

\label{sec:Appendix}

\end{document}